\documentclass[jou]{apa}

\usepackage{amsmath,amssymb,epsfig,multirow}
\usepackage[tight]{subfigure}
\usepackage{subfig}
\usepackage{graphicx}
\usepackage{threeparttable}
\usepackage{array}
\newcolumntype{C}[1]{>{\centering\arraybackslash$}p{#1}<{$}}

\usepackage{setspace}

\def\reals{\mathbb{R}}

\def\sign{\mathop{\rm sign}}

\def\comp{\raise 1pt \hbox{$\scriptstyle\circ$}}

\def\argmax{\mathop{\rm argmax}\limits}
\def\minimize{\mathop{\rm minimize}\limits}

\def\st{\mathop{\rm subject\ to}}

\def\upto{{\raise 1pt \hbox{$\scriptstyle \,\nearrow\,$}}}
\def\downto{{\raise 1pt \hbox{$\scriptstyle \,\searrow\,$}}}

\newtheorem{theorem}{Theorem}

\newtheorem{definition}[theorem]{\bf Definition}
\newtheorem{example}[theorem]{Example}

\newcounter{exhibits}
\rightheader{}
\leftheader{}
\shorttitle{ }
\title{\textbf{Good Debt or Bad Debt: Detecting Semantic Orientations in Economic Texts}}

\fiveauthors{Pekka Malo$^{\star}$}{Ankur Sinha}{Pyry Takala}{Pekka Korhonen}{Jyrki Wallenius}
\fiveaffiliations{\small Department of Information and Service Economy, Aalto University School of Business \\ P.O. Box 21210, FI-00076 AALTO, FINLAND\\ $^{\star}$Corresponding author, e-mail:pekka.malo@aalto.fi \\ tel:+358 40 353 8071, fax: +358 9 431 38535}
{\small Department of Information and Service Economy, Aalto University School of Business \\ P.O. Box 21210, FI-00076 AALTO, FINLAND \\ e-mail: ankur.sinha@aalto.fi \\ tel:+358 40 353 8097, fax: +358 94 313 8535}
{\small Department of Finance and Accounting, Aalto University School of Business \\ P.O. Box 21210, FI-00076 AALTO, FINLAND \\ e-mail: pyry.takala@aalto.fi \\ tel:+358 40 051 0205}
{\small Department of Information and Service Economy, Aalto University School of Business \\ P.O. Box 21210, FI-00076 AALTO, FINLAND \\ e-mail: pekka.korhonen@aalto.fi \\ tel:+358 50 383 6191, fax: +358 94 313 8535}
{\small Department of Information and Service Economy, Aalto University School of Business \\ P.O. Box 21210, FI-00076 AALTO, FINLAND \\ e-mail: jyrki.wallenius@aalto.fi \\ tel:+358 40 353 8100, fax: +358 94 313 8535}

\acknowledgements{This work has been partly supported by Emil Aaltonen Foundation and Marcus Wallenberg Foundation.}

\begin{document}

\maketitle

\begin{abstract}

The use of robo-readers to analyze news texts is an emerging technology trend in computational finance. In recent research, a substantial effort has been invested to develop sophisticated financial polarity-lexicons that can be used to investigate how financial sentiments relate to future company performance. However, based on experience from other fields, where sentiment analysis is commonly applied, it is well-known that the overall semantic orientation of a sentence may differ from the prior polarity of individual words. The objective of this article is to investigate how semantic orientations can be better detected in financial and economic news by accommodating the overall phrase-structure information and domain-specific use of language. Our three main contributions are: (1) establishment of a human-annotated finance phrase-bank, which can be used as benchmark for training and evaluating alternative models; (2) presentation of a technique to enhance financial lexicons with attributes that help to identify expected direction of events that affect overall sentiment; (3) development of a linearized phrase-structure model for detecting contextual semantic orientations in financial and economic news texts. The relevance of the newly added lexicon features and the benefit of using the proposed learning-algorithm are demonstrated in a comparative study against previously used general sentiment models as well as the popular word frequency models used in recent financial studies. The proposed framework is parsimonious and avoids the explosion in feature-space caused by the use of conventional n-gram features.

\end{abstract}

\section{Introduction}

The anxiety surrounding the on-going financial crisis has spurred various studies on market sentiments and how to measure them in a reliable and timely manner. The problem with classical methods, such as confidence indicators computed from regular surveys, is that the information content is strongly restricted by questionnaire design and the results become quickly outdated. To overcome the restrictions with survey-based sentiments, there has been an increasing interest towards investigating techniques that allow investors and officials to monitor economic developments in close to real-time. One promising approach is to consider media as a source of investor sentiments~\cite{mitra11,mitra10,mitra09,dzielinski11,tetlock07,barber08,tetlock08}. Instead of asking for opinions on the recent economic developments in monthly surveys, one can use financial news to approximate the evolution of market sentiments. Since investment information is largely based on news, it is justified to assume that media influences investor sentiment and the behavior of market participants.

Rapid development of natural language processing technologies and access to cheap computational capacity have paved the way for automatic sentiment analysis and the emergence of robo-readers in computational finance. However, the technology is still in its nascent state. Sentiment analysis has turned out to be a complicated and strongly domain-dependent task. Approaches that have proven successful in one domain may not be easily transferred to another one. Not only do the words have different meanings, but the ways in which language is used change by domain. As indicated by Loughran and McDonald's~\citeyear{loughran11} recent study on financial vocabulary, the notion of domain-dependence is particularly strong in a financial context and it requires expert information to be able to define what kind of polarities different words have. 
Given that practically all sentiment analysis techniques rely on word lists as a source of sentiment cues, access to a high quality domain-specific lexicon is a prerequisite for building a successful sentiment model. However, alongside vocabulary related issues, one should also be able to handle domain dependent choices of linguistic structures. Therefore, it is worthwhile to invest in a learning algorithm that is able to handle both aspects of domain-dependence in a flexible manner.

In this paper, we present the Linearized Phrase Structure (LPS) model for detecting semantic orientations in short economic and financial text fragments. The focus is on a phrase or sentence level classification task, where the objective is to categorize any encountered news fragment into a positive, negative or neutral class depending on the tone of the text from an investor's perspective. Although the terms sentiment\footnote{In the literature, sentiment analysis does not refer only to polarity or subjectivity detection, but also on affection. However, in this paper, we have used the term sentiment analysis interchangeably with polarity detection.} and semantic orientation are often used interchangeably in the literature, we find the notion of semantic orientation a better description for the types of sentiments observed in financial texts. In other domains, such as movie or product reviews, the sentiments are often expressed with combinations of adjectives and adverbs, whereas the financial sentiments have often more to do with expected favorable or unfavorable directions of events (e.g., ``the profit is expected to increase") from an investor's perspective. One of the contributions is the enrichment of the existing lexicons with economic concepts and information on how their semantic orientation varies by context. Although many of the financial concepts are neutral in nature, they may still lead to a strong semantic orientation when combined with verbs and directional information. The advantage of the LPS model is in its ability to account for important interactions between relevant financial concepts and other phrase-structure components. The algorithm extends the quasi-compositional polarity-sequence framework presented by Moilanen, Pulman and Zhang~\citeyear{moilanen10} to accommodate the interactions which are typical for phrase-structures encountered in financial and economic texts. As observed from the results of a comparative study, the interactions and context of semantic orientations plays an important role in phrase-level analysis. A good lexicon provides a solid starting-point, but the performance can be substantially improved by a learning algorithm that can capture the domain-specific use of language. 

The article is organized as follows. In the next section, we discuss the benefits of machine learning algorithms for sentiment analysis and our contributions in relation to the recent work done within the financial domain. The third section provides a description of our domain-specific financial lexicon and motivates the newly included features. This is followed by the presentation of the Linearized Phrase Structure (LPS) model for detecting semantic orientations.  In addition to the contributions towards algorithm and lexicon development, we present a new open-source phrase-bank for training and evaluating models for financial and economic news texts. The dataset provides a collection of $\sim5000$ phrases/sentences sampled from financial news texts and company press releases, which are tagged as positive, negative or neutral by a group of 16 annotators with adequate business education background. Thereafter, in the results and experiments section, we perform a comparative study with a number of baseline approaches for sentiment analysis. An error analysis of the LPS model is also provided along with discussions for future developments.

\section{Related work and contributions}

Sentiment analysis is a multidisciplinary field, which has greatly benefited from the recent advances in statistical natural language processing and machine learning~\cite{thelwall10,thelwall12,moilanen10,moilanen07,pang05,wilson09}. Most of the recent work has focused on the analysis of opinionated blog-writings and informal texts in social media~\cite{young12,mitrovic11,bollen11,balahur09,thelwall12}. There are only few studies that have investigated how sentiments should be modeled in financial and economic domains. 

\subsection{Recent developments in financial domain}

Previous research on textual analysis in a financial context has primarily relied on the use of word categorization (``bag of words") methods to measure tone~\cite{tetlock07,tetlock08,engelberg08}. In particular, there has been interest towards understanding how negative vocabulary in firm-specific earnings news stories could be utilized in forecasting future performance. Perhaps the most prominent effort to advance the development of sentiment analysis techniques for finance has been the recent introduction of a new polarity lexicon by Loughran and McDonald~\citeyear{loughran11}. The authors perform a careful study to examine the quality of word classifications given by the Harvard Dictionary. As a result of the study, they make a note of commonly misclassified terms and propose a revised dictionary, which is better fitted for financial texts. Loughran and McDonald also propose a few other word classifications (positive, uncertain, litigious, strong modal and weak modal), which should be considered along with lists of negative words. The studies pre-dating the paper have almost exclusively relied on the approach suggested by Tetlock~\citeyear{tetlock07}, who was among the first to demonstrate the benefits of using General Inquirer in a financial context. It is also quite common to adopt an off-the-shelf package such as Diction or Wordstat to perform dictionary-based searches for sentiment cues~\cite{demers08,davis07}. 

To the best of our knowledge, only a handful of studies have chosen to use statistical or machine learning techniques for financial sentiment analysis~\cite{antweiler04,das07,li09,ohare09}. Li~\citeyear{li09} uses a Na\"{i}ve Bayesian classifier to perform a bag-of-words based classification without the help of a dictionary. Also O'Hare et al.~\citeyear{ohare09} use a bag-of-words representation to train Na\"{i}ve Bayes and SVM classifiers, but unlike Li~\citeyear{li09} they focus on blogs and perform the classification on a document-topic level instead of sentences. Das and Chen~\citeyear{das07} resort to a more sophisticated approach of combining a dictionary-based approach with a complicated multiple-classifier voting system, which combines word count based classifiers with Bayesian and Discriminant-based classifiers. However, none of the approaches makes a clear effort to develop a model that can utilize the domain-dependent use of language. Though lexicons help to pick up the vocabulary that provides cues on sentiments, the overall sentiment of phrases is still strongly dependent on the way the phrases have been structured. As indicated in the recent study on general sentiment analysis by Wilson et al.~\citeyear{wilson09}, the contextual polarity of the phrase in which a particular instance of a word is detected may differ from the word's prior polarity suggested by a lexicon. Therefore, when building a model to capture semantic orientations, one should not overlook the contextual cues available in the phrase-structures. The arguments presented by Wilson et al.~\citeyear{wilson09} are quite convincing.

\subsection{Contributions} 

The contributions of the paper are summarized as follows.

\subsubsection{Establishment of a phrase-bank for financial and economic news texts}

The key arguments for the low utilization of statistical techniques in financial sentiment analysis have been the difficulty of implementation for practical applications and the lack of high quality training data for building such models.\footnote{Except for a few special cases~\cite{mishne05,mishne06} human-coded datasets are practically always necessary when applying machine-learning or statistical techniques.} Especially in the case of finance and economic texts, annotated collections are a scarce resource and many are reserved for proprietary use only~\cite{ohare09}. To resolve the missing training data problem, we present a collection of $\sim 5000$ sentences to establish human-annotated standards for benchmarking alternative modeling techniques. The annotators were carefully screened to ensure that they have sufficient business knowledge and educational background.

\subsubsection{Introduction of verbs and directional expressions to a finance lexicon}\label{sec:contrib2}

In addition to the use of negative word lists, there are also many other word categories that can be helpful in detecting semantic orientations in financial and economic texts. Using the more general MPQA lexicon by Wiebe et al.~\citeyear{wiebe05} and the financial polarity-lexicon by Loughran and McDonald~\citeyear{loughran11} as a starting point, we propose the following modifications that infuse further domain-specific knowledge into the sentiment models: (1) addition of domain-specific concepts which can influence the overall semantic orientation of a sentence; (2) addition of verbs and expressions which help to detect the direction of events (e.g., whether the profit is expected to increase or decrease); (3) addition of information on how the polarity of different concepts depends on the expected direction of events (e.g., result is positive when it is expected to increase, but neutral or negative when declining).

\subsubsection{Linear Phrase Structure (LPS) model for semantic orientations}

To benefit from the domain-specific knowledge, which we have proposed to add into the finance lexicons, one needs to have a model that is not restricted to frequencies of positive or negative words but is able to take the entire phrase-structure into account. Further, to avoid the curse of dimensionality, it is wise to be selective in how the sentence structure information is incorporated into a model. A recent paper by Moilanen et al.~\citeyear{moilanen10} examines a number of ways to use syntactic information in sentiment models without the inclusion of any conventional n-gram features. An important conclusion of their study is that not all syntactic information is relevant for phrase-level sentiments. In fact, their most robust model is a simple polarity-sequence framework where the phrase-structure of a sentence can be reduced to a sequence of slices, where each slice carries information about the polarity (i.e. positive, negative, neutral) of a small part of the sentence. In this paper, we extend the polarity-sequence framework of Moilanen et al.~\citeyear{moilanen10} by accommodating elements which are particularly relevant for the financial domain. The most important differences are: (1) the inclusion of finance-specific entities into the polarity sequence model; and (2) the inclusion of interactions between financial concepts and verbs or other direction-giving expressions (for further details, see the next section). The model is relatively simple to implement, and it is not strongly dependent on deep phrase-structure parsing.

\section{Semantic orientations in economic texts}\label{sec:orientations}

The use of polarity lexicons is perhaps the most commonly adopted approach for identifying semantic orientations (i.e. positive, negative or neutral) in text. However, it is well recognized that the performance of approaches based on polarity-lexicons is largely affected by the domain and context of the subjective or opinionated statement. An expression that can be considered to have a strong semantic orientation in one domain may need to be interpreted differently in another context. In this section, we elaborate aspects that should be taken into account when building a domain-specific lexicon for analyzing economic and financial texts. In addition to the requirement for being able to account for general polarity-bearing expressions correctly, we consider how phrase-structure information and contextual knowledge can be utilized to improve the accuracy of semantic orientation judgments. In particular, we investigate how ontological knowledge on inherent semantic-orientations of financial concepts and their dependencies on directional-expressions help to detect financial sentiments.

\subsection{Domain-adjusted lexicon entries}\label{sec:lexicon-entries}

For the experiments carried out in this paper, we utilize a domain-specific lexicon with over 10,000 entries. The entries can be classified into the following categories depending on their roles: (i) general expressions with a semantic orientation; (ii) financial entities; (iii) entities which influence the semantic orientation of other entities (e.g., negators, booster words, expressions describing change or direction of events, and entities that describe a degree of uncertainty). A summary of the number of different lexicon entities is provided in Table~\ref{tab:lexicon-stats}.

\begin{table}[hbt]
\caption{Financial lexicon statistics}
\label{tab:lexicon-stats}
\begin{center}
\begin{small}
\begin{tabular}{l|c|c|c} \hline
\hline
Entity class	&	Category	&	Number	&	\% of all entities	\\
\hline
General entity	&	Positive	&	2933	&	27.8	\\
	&	Negative	&	5951	&	56.3	\\
	&	Neutral	&	585	&	5.5	\\
\hline
Financial entity	&	Positive-if-up	&	252	&	2.4 \\
	&	Negative-if-up	&	95	&	0.9 \\
\hline
Direction	&	Down	&	128	&	1.2 	\\
	&	Up	&	186	&	1.8 \\
\hline
Polarity influencer	&	Reversal	&	188	&	1.8 \\
	&	Modal	&	50	&	0.5 \\
	&	Litiguous	&	95	&	0.9 	\\
	&	Uncertain	&	102	&	1.0 	\\
\hline
\end{tabular}
\end{small}
\end{center}
\end{table}


\subsubsection{General expressions with polarity}

The first group of lexicon entries consists of commonly encountered opinionated or subjective expressions. In our paper, the definition of these entities builds on the lexicon derived from the Multi-perspective Question Answering (MPQA) corpus of opinion annotations~\cite{wiebe05,wilson08,wilson09}.  The MPQA lexicon consists of only single word clues, which were compiled by merging the lists of negative and positive words from Riloff and Wiebe~\citeyear{riloff03}, Hatzivassiloglou and McKeown~\citeyear{hatzivassiloglou97}, and General Inquirer by Stone, Dunphy, Smith and Ogilvie~\citeyear{stone66}. Each entry is equipped with information on degree of subjectivity, prior-polarity, part-of-speech and lemma. To accommodate the domain specific requirements, we augmented the MPQA-based dictionary with the finance-specific lists compiled by Loughran and McDonald~\citeyear{loughran11}. When overlaps were encountered while merging the lists, the financial domain sentiment was preferred over general prior-polarities specified by MPQA.

\subsubsection{Financial entities}

Inference of the semantic orientation of an entity based on its association with other entities can be quite challenging, especially when the semantic orientation of an expression is largely modified by the other constituents of the phrase structure. For example, a simple phrase ``profit was 10 million" is a neutral statement, whereas ``profit increased dramatically from last year" has a positive semantic orientation. One way to take these aspects into account is to incorporate further human knowledge into the lexicon. As a basis for this task, we consider a specialized financial dictionary that provides information on the types of concepts and how their semantic orientation is modified by the direction of events (e.g., increasing profit vs. decreasing profit). 

\begin{figure}[ht]
\begin{center}
\includegraphics[scale=0.37,angle=90]{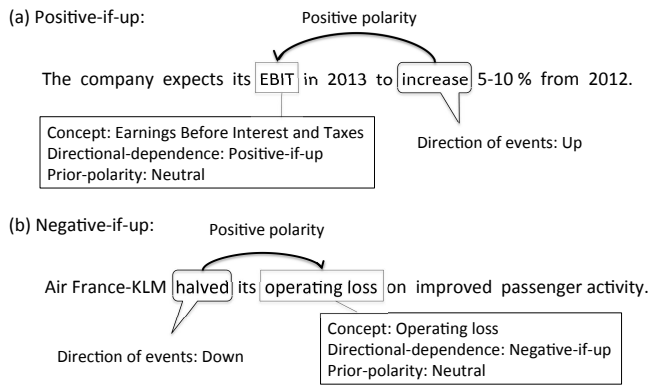}
\vspace{2mm}
\caption{Directional-dependence: Semantic orientation of a financial entity often depends on the favorable or unfavorable direction of events.}
\label{fig:directional-dependence}
\end{center}
\end{figure}

The relevant domain knowledge is augmented into the lexicon by introducing a new entity-type, {\it financial entities}, which covers a broad range of commonly encountered financial concepts.\footnote{Whereas our original collection of financial entities extracted from Investopedia featured more than 16,000 terms, we pruned the set down to 684 central concepts, which are most commonly encountered in financial texts.  To zero in on the most important terms, we took a random sample of 100,000 news articles and counted the occurrences of all financial terms in the news. Out of the sample of 684 terms that we manually went through, 51 terms have a very clear effect on sentiment, and 177 have an effect on sentiment in most cases when combined with a verb representing movement up or down.} Each financial entity is expected to carry the following information:
\begin{itemize}
\item[(i)] Concept: Description of the underlying unique financial concept to which the lexicon entry refers (e.g., Earnings Per Share).
\item[(ii)] Anchor text: Expression used as a clue for finding the given financial concept in text (e.g., EPS).
\item[(iii)] Prior-polarity: For most of the financial entities, the default semantic orientation is neutral. In general, the polarity of a financial entity is defined conditionally depending on whether it is moving up or down.
\item[(iv)] Directional-dependence: Describes the semantic orientation of the concept when it is modified by other parts of the phrase, indicating a direction of events, i.e. what is the polarity of the concept if it is increasing or decreasing; see Figure~\ref{fig:directional-dependence}. The financial entities included in the lexicon are one of the following types: positive-if-up or negative-if-up. For example, we classify EBIT (earnings before interest and taxes) as a positive-if-up concept and liability as a negative-if-up concept. A concept is of type positive-if-up if its semantic orientation becomes positive when it is increasing (up). Similarly, a concept is treated as negative-if-up, if its semantic orientation becomes negative when it is increasing (up).
\end{itemize}

\subsubsection{Directionalities and other polarity influencers}

As discussed by Wilson et al.~\citeyear{wilson09}, phrase-level sentiment analysis is a complicated task which cannot be solved by detecting prior-polarities only. In addition to prior-polarities, there are a number of other factors that can influence the overall semantic orientation of a phrase. This broad class of operators, which can modify the contextual semantic orientation, is generally referred to as {\it polarity influencers}, also known as lexical valence shifters or degree modifiers~\cite{wilson09,polanyi04}. The most commonly encountered polarity influencers tend to fall into one of the following categories: (i) negators; (ii) boosters and diminishers; and (iii) modal operators. 

Motivated by our earlier discussion on the semantic orientations of financial entities (see previous subsection on financial entities), we have extended the above list of commonly encountered polarity influencers by adding the following class of polarity influencers, which play an important role in economic and financial texts: the directionalities.  Since the semantic orientation associated with a financial entity may depend on whether it is going up or down (e.g., increasing loss vs. decreasing loss), we define directionality as a word or n-gram describing change or direction of events. Examples of simple directionalities are verbs such as {\it increase, decrease, balloon, rocket} and {\it plummet}. The directionalities operate in a similar manner as boosters and diminishers, but in addition to modifying the degree of the orientation they decide the type of the semantic orientation as well.  For example, when considered out of context, the statement ``Profit of the accounting period was EUR 0.3 mn" is neutral, but ``Profit {\it fell} by 33\% from the third quarter" has a negative orientation.  

As a seed for creating the list of directionalities, we used the Harvard IV word lists from the General Inquirer. Based on the Harvard categories, two word lists were formed: one for down terms (categories decrease and fall) and another for up terms (categories increase and rise). The lists of candidate directionalities were then further pruned to remove words that have a different meaning in a financial context (e.g., terms such as ÒinflationÓ, ÒdiscountÓ, and ÒrecessionÓ were removed from the lists, since they are better interpreted as financial entities themselves rather than directionalities). The outcome was a list of over 200 directionalities, which was further enriched by adding a few more recently developed expressions.

\subsection{Entity detection and pruning}\label{sec:lexicon-rules}

In this paper, the term phrase-structure is used interchangeably with sentence parse~\cite{chomsky57,manning03}. The information obtained from the syntactic analysis is utilized in two steps: (i) the first pass is an entity identification step: and (ii) the second pass is an entity pruning and combination step, where we can remove neutral entities, identify the directionalities associated with the financial entities, and incorporate the effects of polarity influencers.

\subsubsection{Entity detection rules}

To ensure that a candidate expression will most likely have the same meaning as indicated by a corresponding lexicon entry, a number of properties can be considered to reduce the number of false positives. In this paper, we have used a combination of part-of-speech (POS) tag based rules for unigrams as well as phrase structure information when the expression at hand is an n-gram. For most entities, the phrase level rules can be viewed as matching of ordered sequences of POS tags.

\subsubsection{Entity pruning rules}

\begin{figure*}[ht]
\begin{center}
\includegraphics[scale=0.45]{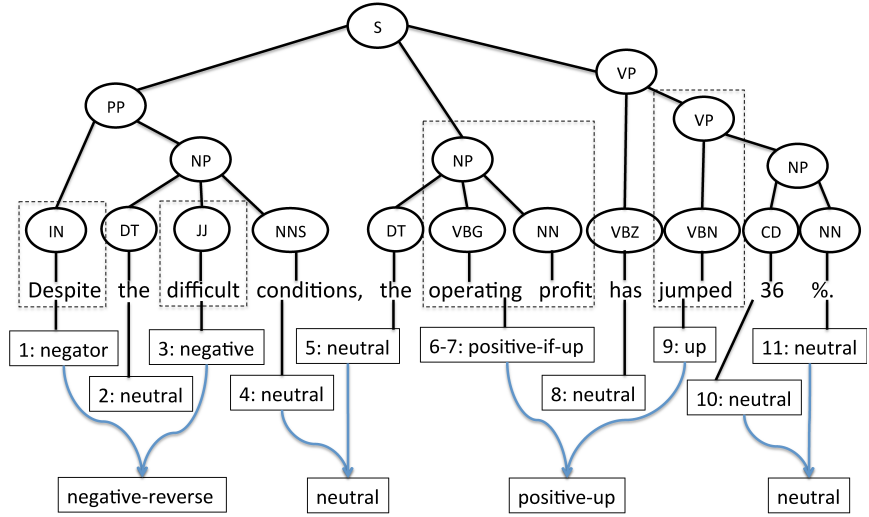}
\vspace{5mm}
\caption{Entity pruning rules: To reduce the number of entities and improve the information value of the remaining ones, heuristic rules are applied to merge neutral entities and take into account the effects of polarity influencers on other entities.}
\label{fig:entity-pruning}
\end{center}
\end{figure*}

Once the initial set of entities has been recognized, neutral entities are merged before taking into account the effects of polarity influencers on the semantic orientations of other entities:
\begin{itemize}
\item[(i)] Merge-neutrals-rule: If several neutral entities occur in a sequence, they can be combined into a single neutral entity which spans a large part of the given phrase. 
\item[(ii)] Polarity-influence-rule: When a phrase contains both a polarity influencer and another entity whose polarity is modified by the influencer, we apply the influencer directly and retain only the main entity with modified polarity. For example, consider the two phrases in Figure~\ref{fig:directional-dependence}, where in both cases we find a directionality which modifies the polarity of a financial concept. Instead of retaining both entities, we combine the entities by retaining financial concept and adjusting its prior-polarity from ``neutral" to a modified polarity ``positive-up" or ``negative-up" reflecting the fact that polarity depends on the direction of events. The impacts of other influencers are accounted in similar manner; see Figure~\ref{fig:entity-pruning}.  
\end{itemize}

When applying the pruning rules, we do not hide the impact of polarity influencer. Instead of using the main semantic orientations ``positive, neutral, negative", we distinguish the impact of pruning by adding postfix to the prior-polarity of the entity (e.g., ``positive-up" instead of ``positive"). 

\section{Linearized Phrase-Structure Model}\label{sec:lps-model}

In this section, we present the Linearized Phrase-Structure (LPS) Model for predicting semantic orientations of short economic texts. As described in the previous section, in order to operate in financial and economic domains the framework should be able to recognize financial domain concepts and identify their semantic orientation based on phrase structure information and domain knowledge. In addition to the domain specific aspects, the model should also have an ability to resolve conflicting cases where a phrase contains several  semantic orientations, and be easy to retrain based on the feedback given by the users. To accommodate these requirements, the LPS-model is constructed in three stages: (i) extraction of entities with semantic orientation; (ii) phrase structure projection step; and (iii) multi-label classification step. 

In a nutshell, the model works as follows. When an incoming phrase is picked up from an economic news stream, we first recover the phrase-structure using statistical techniques. Then the phrase structure is quickly traversed by a sequence of entity-recognizers, which try to locate the parts of structure corresponding to one of the lexicon entity types described in the previous section. Once the phrase structure has been transformed into a sequence of lexicon entities, we project the entities into a sequence in $l^2$-space. Each projected sequence has an interpretation as a representative of an equivalence class of phrases with similar features, and their length varies depending on the complexity of the underlying phrase structure. The obtained $l^2$-sequences are then presented to a linear multi-label classifier, which learns to associate the sequences with corresponding semantic orientations indicated by the user. 

\subsection{Extraction of entities with semantic orientation}\label{sec:sos}

From modeling perspective, the lexicon entities are considered to be the smallest units with a perceptible semantic orientation. Each phrase can contain one or more lexicon entities, which jointly provide the information needed to determine the overall semantic orientation of the phrase. Formally, we can write the entity-extractor mapping as follows.

\begin{figure*}[htbp]
\begin{center}
\includegraphics[scale=0.5]{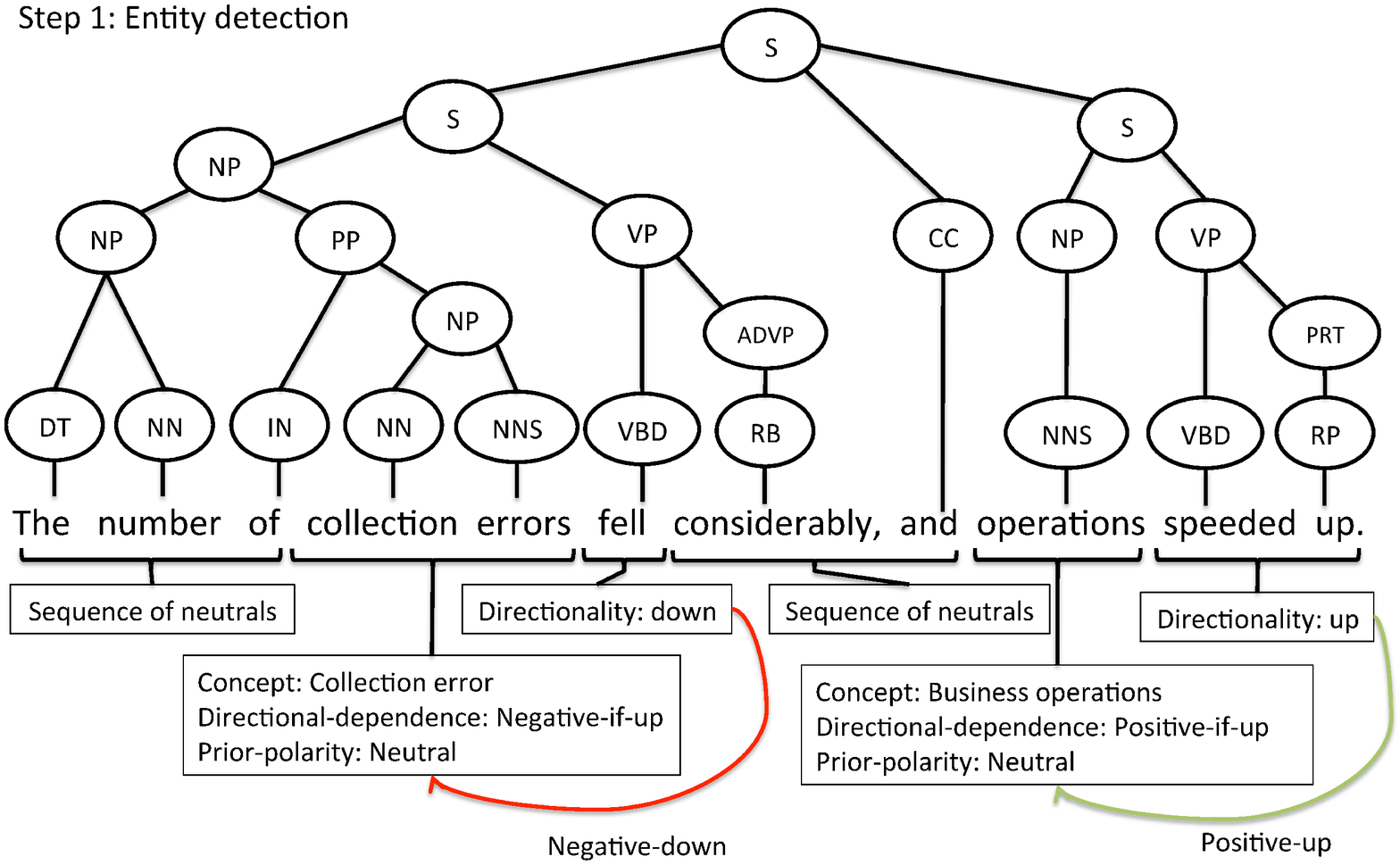}
\vskip 1cm
\includegraphics[scale=0.5]{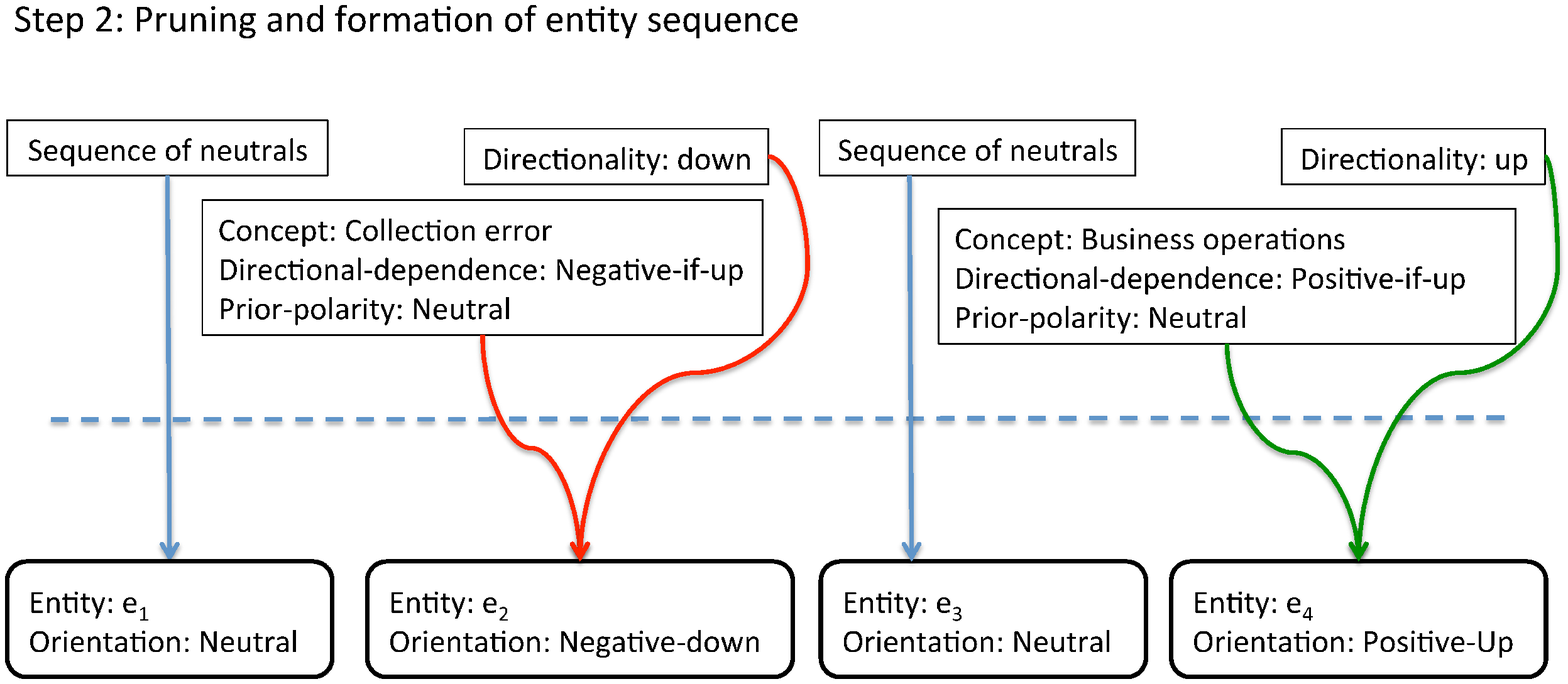}
\vspace{5mm}
\caption{The entity extractor is operationalized in two steps: (1) The first step is to identify all entity candidates: For every lexicon entry, we check whether a subset of leaves in the phrase-structure matches; (2) The second step is to replace the leaves with lexicon entities, and apply pruning rules to merge neutral entities and take into account the effects of polarity influencers. As final output, the extractor maps the phrase into a sequence of entities, which are positioned in their order of appearance.}
\label{fig:entity-extraction}
\end{center}
\end{figure*}


\begin{definition}[Entity extractor]\label{def:entity-extractor}
Let $E$ denote the collection of lexicon entries, and let $S$ be the space of possible phrase-structures. An entity-extractor is defined as a mapping $e: s \mapsto e_1 e_2 \dots e_n$, which presents a given phrase-structure $s$ as a sequence of matched lexicon-entities, $e_i\in E$, $i\in\{1,\dots,n\}$. The entities are provided according to their location in the original phrase. If several entities match the same part of the phrase, then the entity with longest span of the phrase is chosen.
\end{definition}

\begin{figure}[h]
\centering
\includegraphics[scale=0.5]{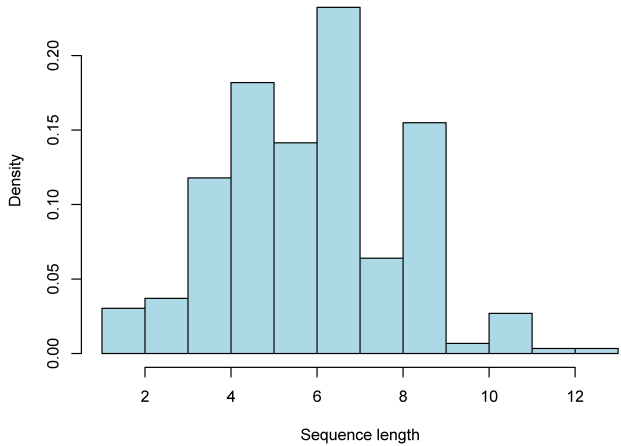}
\vspace{5mm}
\caption{The figure shows the distribution of entity-sequence lengths (before pruning) in the financial phrase-bank, which consists of ~5000 sentences. An entity-sequence is a linear representation of the phrase structure produced by the entity-extractor.}\label{fig:pattern-length}
\end{figure}

The entity extractor can be operationalized as follows: (1) The first step is an entity identification problem: For every lexicon entry $e_i$, we check whether a subset of the leaves matches with the entry. If a candidate is found, we traverse the phrase-structure tree to see that all matching rules are met; (2) The second step is to replace the leaves with lexicon entities: If matching criteria were met, then we associate entity $e_i$ with the position of the matched leafs. At this stage we also apply the entity pruning rules to merge neutral entities and take into account the effects of polarity influencers as discussed in the previous section. The leaves that do not associate with any of the lexicon entities are associated with general neutral entities spanning the corresponding positions. As a result, the phrase $s$ can be represented as a sequence of entities, which are positioned in their order of appearance. See Figure~\ref{fig:pattern-length} for the distribution of entity-sequence lengths in our financial news phrase bank before applying the pruning rules.

\begin{example}
Consider, for example, the phrase ``The number of collection errors fell considerably, and operations speeded up"  in Figure~\ref{fig:entity-extraction}. As an outcome of the entity detection step, a list of candidates is produced. In the figure, the spans of the entity candidates have been marked with brackets and a description of each entity is provided in a box. To simplify the picture, we have also grouped neutral expressions (1-3; 7-8) together and used colored arrows to show the situations were polarity influencers are at work. In the second step, pruning rules are applied to merge suitable entities together. 

%
%
%

\end{example}

When implementing the entity extractor, several computational simplifications can be proposed. In particular, when the use of a full phrase structure parser is considered to be computationally infeasible, it is still possible to replicate the pruning-rules using a high quality part-of-speech tagger, since the entity identification step can be carried out with good accuracy by relying on part-of-speech tags alone also for n-grams. The only difference is that instead of defining the pruning rules using the phrase-structure parse tree, we can consider the raw sequence obtained from Step 1 and perform entity merging within a restricted window around each entity, which may be affected by a polarity influencer. For instance, in the above example, the choice of window length 1 would produce correctly behaving pruning rules, since the neighborhoods of entity 4 and entity 8 both contain a corresponding polarity influencer. Consequently, one may suffer only a minor loss of accuracy in the case of complicated sentences, but this is likely to be offset by the corresponding gain in computational speed.

\subsection{Phrase Structure Projection}\label{sec:psp}

Assuming that the phrase structure combined with lexicon's contextual knowledge provides enough cues to discriminate between different semantic orientations, it makes sense to glue together all such sentences, whose phrase structures feature similarly typed lexicon-entities in the same order. With this purpose in mind, we consider the following way to recognize whether phrase structures are equivalent from the perspective of analyzing their semantic orientations.

\begin{definition}[Equivalence of phrase structures]\label{def:equiv}
Let $e$ be the entity extractor given by Definition~\ref{def:entity-extractor}. A pair of phrase-structures $s_1, s_2 \in S$, with entity sequences $e(s_1)= e_{11} e_{12} \dots e_{1n}$ and $e(s_2)= e_{21} e_{22} \dots e_{2m}$, are said to be equivalent from the perspective of semantic orientations when their entity-sequences are of equal length and the entities are of the same type, i.e.  if $n=m$ and type of $e_{1i}$ is the same as type of $e_{2i}$ for every $i\in\{1,\dots n\}$.
\end{definition}


Using the entity sequence based approach to define an equivalence relation on the space of phrase structures, we can 
now present phrase structure projection as a mapping which bundles together all phrases that can be considered to be structurally similar from the perspective of analyzing their semantic orientations.

\begin{definition}[Phrase Structure Projection]\label{def:psp}
Let $S$ be the space of phrase structures. The choice of structural equivalence relation $\theta$ on $S$ can be used to define a quotient mapping from phrases to their corresponding equivalence classes, i.e.
$$
\varphi_{\theta}: S \to S/{\theta}
$$
where $S/{\theta}$ denotes the partitioning of the phrase structure space into classes of sentences with similar structural features. For each phrase $s$, the corresponding equivalence class $\varphi_{\theta}(s) \in S/{\theta}$ has a unique representation as a bit-sequence in $l^2$-space.
\end{definition}

In practice, given a phrase $s$, construction of the bit sequence presentation for $\varphi_{\theta}(s)$ is easy once the entity sequence $e(s)$ is known. One simply needs to agree on the appropriate coding for each entity type. In this paper, we have chosen to write $\varphi_{\theta}: s \mapsto \tilde{e}_1\dots \tilde{e}_n$ where $\tilde{e}_i\in \{0,1\}^{m}$ is an indicator for the type of the matched entity $i$ and $m$ is the number of entity types in the lexicon. In this construction, the number of entities $n$ is always less than the number of leaves in the phrase structure, which implies that the length of a bit sequence representation for $\varphi_{\theta}(s)$ is always bounded by the number of leaves times the number of alternative entity types. The representation is also able to scale automatically according to the complexity of the phrase structures, and we avoid making ad-hoc cut-offs on the size of the bit sequence presentation. 


\subsection{Learning mechanism}\label{sec:learn}

Given a collection of human-annotated text segments, the remaining problem is to learn a mapping that associates the linearized constituency representations of sentences with correct semantic-orientations. However, there are two essential requirements, which the learning mechanism should satisfy: (i) ability to handle large-dimensional feature spaces in an effective manner; and (ii) ability to perform multi-label classification. As noted in the previous section, the dimension of the linearized constituency projections is unlimited by default, and the obtained feature sequences are allowed to scale depending on the length of the input text segments. Even for relatively short sentences, the dimension of the projection can grow quite large. 

To resolve these challenges, the chosen learning mechanism should be robust for large dimensional feature spaces and have an inherent ability to ignore irrelevant features. Out of the alternative techniques with a proven track record for handling polychotomous classification problems, the family of SVM models is most suitable for our purposes. After a few preliminary experiments, we decided to choose the following multi-class SVM approach with ``one-against-one" strategy which has shown good performance in comparison to alternatives based on ``all-together" methods or ``one-against-all" and DAGSVM strategies. The choice is also well supported by the study of Hsu and Lin~\citeyear{hsu02}, who evaluated a number of alternative multi-class SVM models with different estimation strategies in the light of large-scale problems.

\begin{definition}[Semantic-orientation mapping]\label{def:learn}
Let $\mathcal{H}_{\theta}$ denote the hypothesis space of alternative semantic-orientation mappings, $h_{\theta}:S\to P$, and let $T=\{(s_t,p_t)\}_{t=1}^n$ be a training sample of human annotated text segments.
 In accordance with the pairwise multilabel classification principle, the optimal semantic-orientation mapping is then given by
$$
h_{\theta}(s)=\argmax_{i=1,\dots,k}h^i(s), \quad h^i(s)=\sum_{j\neq i, j=1}^k \sign(h^{ij}(s))
$$
where each mapping $h^{ij}(s)=\langle w^{ij},\varphi_{\theta}(s)\rangle+b^{ij}$ is a decision function for class $i$ against class $j$, and $\langle \cdot , \cdot \rangle$ denotes the inner product of two possibly infinite dimensional vectors in the $l^2$-space. The parameters $w^{ij}$ and $b^{ij}$ are obtained by solving the following binary classification problem:
\begin{align*}
\minimize_{w^{ij},b^{ij},\xi^{ij}}\quad & \frac{1}{2}\langle w^{ij},w^{ij}\rangle + C\sum_{t=1}^n \xi_t^{ij}\\
\st\quad  & \langle w^{ij}, \varphi_{\theta}(s_t)\rangle+b^{ij}\geq 1-\xi_t^{ij}, \quad\text{if $y_t=i$},\\
 & \langle w^{ij},\varphi_{\theta}(s_t)\rangle+b^{ij}\leq -1+\xi_t^{ij}, \quad\text{if $y_t=j$}, \\
 & \xi_t^{ij}\geq 0,
\end{align*}
where $\varphi_{\theta}$ is the phrase-structure projection in Definition~\ref{def:psp}.
\end{definition}
In practice, the $k(k-1)/2$ estimation problems required to obtain the semantic-orientation mapping can be efficiently solved by considering the dual formulation with kernel given by $K:S\times S \to \reals$, $K(s_i,s_j)=\langle \varphi_{\theta}(s_i),\varphi_{\theta}(s_j)\rangle$.

\section{Annotating financial news}\label{sec:phrase-bank}

As discussed by Loughran and McDonald~\citeyear{loughran11}, it is well understood that the vocabulary and expressions used to describe economic events and company related news are not identical across media. Until now, very limited efforts have been taken to build corpora that cover economic or financial domains. To the best of our knowledge none of the existing datasets provides phrase-level annotations for news documents. To alleviate this data gap\footnote{In order to foster further research on financial sentiment analysis, the dataset is made available for research purposes under appropriate license.}, we will now briefly outline our recently built financial news phrase dataset, which is used as a gold standard for evaluating the performance of  sentiment models dedicated for economic texts. 

\subsection{Building the phrase bank}

The corpus used in this paper is made out of English news on all listed companies in OMX Helsinki. The news has been downloaded from the LexisNexis database using an automated web scraper. Out of this news database, a random subset of 10,000 articles was selected to obtain good coverage across small and large companies, companies in different industries, as well as different news sources. Following the approach taken by Maks and Vossen~\citeyear{maks10}, we excluded all sentences which did not contain any of the lexicon entities. This reduced the overall sample to 53,400 sentences, where each has at least one or more recognized lexicon entity. The sentences were then classified according to the types of entity sequences detected. Finally, a random sample of $\sim5000$ sentences was chosen to represent the overall news database.

\subsection{Annotation task and instructions}

The objective of the phrase level annotation task is to classify each example sentence into a positive, negative or neutral category by considering only the information explicitly available in the given sentence. Since the study is focused only on financial and economic domains, the annotators were asked to consider the sentences from the view point of an investor only; i.e. whether the news may have positive, negative or neutral influence on the stock price. As a result, sentences which have a sentiment that is not relevant from an economic or financial perspective are considered neutral.

The selected collection of phrases was annotated by 16 people with adequate background knowledge on financial markets. Three of the annotators were researchers and the remaining 13 annotators were master's students at Aalto University School of Business with majors primarily in finance, accounting, and economics. Each annotator was allocated a random subset of sentences and was given 1 month to perform the task. To alleviate the cognitive load, the data was split into small parts over the annotation period. The average number of annotations done by one person was 1500 sentences. Consequently, each sentence of the phrase bank was annotated by 5 to 8 participants. Acknowledging the difficulties involved in phrase-level annotation task, the main task was preceded by a training phase of pilot annotations to ensure that the annotators had a sufficient understanding of the objectives. Similar to Wiebe et al.~\citeyear{wiebe05}, the annotators were instructed to consider the following main guidelines while annotating the phrases:
\begin{itemize}
\item There are no fixed rules about how particular words should be annotated. The instructions provide a handful of examples, but do not state any specific rules.
\item Avoid speculation based on prior knowledge. The annotators should not draw conclusions that require speculation or prior knowledge on the company. Each piece of news should be judged in isolation by using the information that is explicitly available in the given example phrase.
\item The annotators should be as consistent as they can be with respect to their own annotations, and the sample annotations given to them for training.
\end{itemize}

We do not annotate for: whose opinion is at hand, or relevance of the sentence, as we assume that the aforementioned will play a smaller role with sentiment. Furthermore, we assume that most sentences should be relevant for the company for two reasons: first, they are parts of articles that we classified previously as relevant to the company, second: they are mainly texts from financial press that should be concise and to the point.

\subsection{Analysis of the phrase bank}

To measure the reliability of the annotation scheme and examine the degree of agreement between the annotators, we conducted an agreement study with all 16 annotators using a subset of 150 sentences from the phrase-bank. The results of the experiment are summarized in Table~\ref{tab:interannotator}. 

Panel A reports statistics on pairwise-agreements, which are computed as averages across all annotators. The overall average pairwise agreement is 74.9\%, which is quite good considering the limited amount of contextual information available in phrase-level annotation tasks and the large number of annotators. When examining category specific agreements, we find a very high agreement (98.7\%) for separating positive and negative sentences from each other. There is also a strong consensus (94.2\%) for distinguishing neutral sentences from negative ones. However, separating positive sentences from neutral sentences appears to be more challenging, which is reflected as a lower average pairwise-agreement (75.2\%). This finding is somewhat predictable, since making a difference between commonly used company glitter and actual positive statements is not an easy job. Quite often the disagreement on positive vs. neutral cases follows from borderline cases, which could be tagged by human as either/or.\footnote{Consider, for example, the sentence: ``The long-standing partnership and commitment enable both parties to develop their respective operations, and ESL Shipping will also have the opportunity to update its fleet and improve its efficiency."  Although, the sentence suggests that the company's efficiency will increase, a human annotator may well think that the sentence is mainly an example of commonly used company glitter and the increase is not substantial enough.} It is also possible that a higher degree of agreement could have been achieved by having separate categories for moderately positive or negative phrases. 

The multirater agreement statistics are shown in Panel B. When considering the results as a whole with the intra-class correlation coefficient (ICC)~\cite{shrout79} and other reliability measures~\cite{finn70,robinson57} in the range $0.649\sim0.818$, we conclude that the annotations were done with reasonable consistency and reliability. In order to safely utilize the labels from different annotators, we have used a majority vote to construct the reference datasets for experiments.

\begin{table}[hbt]
\caption{Interannotator-agreement statistics based on a subset of 150 sentences tagged by all 16 annotators. ICC denotes the intra-class correlation coefficient. Both consistency and agreement versions of the ICC ratio are reported. Also Finn-coefficient  for reliability and Robinson's A measure for agreement are given.}
\label{tab:interannotator}
\begin{small}
\begin{center}
\begin{tabular}{|l|c|}
\multicolumn{2}{l}{Panel A: Average pairwise annotator agreement} \\
\hline
Number of sentences	&	150	\\
Number of annotators	&	16	\\
Overall agreement	&	0.749	\\
Positive vs. Negative	&	0.987	\\
Negative vs. Neutral	&	0.942	\\
Positive vs. Neutral	&	0.752	\\
\hline
\multicolumn{2}{l}{} \\
\multicolumn{2}{l}{Panel B: Multirater-agreement, reliability and consistency} \\
\hline
ICC consistency	&	0.667	\\
ICC agreement	&	0.649	\\
Robinson's A	&	0.688	\\
Finn-coefficient	&	0.818	\\
\hline

\end{tabular}
\end{center}
\end{small}
\end{table}


\subsection{Definition of the Gold Standard}\label{sec:datasets}

Given the large number of overlapping annotations (5 to 8 annotations per sentence), there are several ways to define a majority vote based gold standard. To provide an objective comparison, we have formed 4 alternative reference datasets based on the strength of majority agreement: (i) sentences with 100\% agreement; (ii) sentences with more than $75\%$ agreement; (iii) sentences with more than 66\% agreement; and (iv) sentences with more than 50\% agreement (simple majority based Gold standard). In the experiments, all datasets have been used to evaluate the performance of the LPS models with respect to a number of competing baseline models. The number of sentences and distribution of labels in each reference dataset are shown in Table~\ref{tab:label-distribution}. All of the datasets are quite similar in terms of label distribution. 

\begin{table*}
\caption{Distribution of labels in phrase-bank for 4 subsets formed based on the strength of majority agreement. Each sentence has 5 to 8 overlapping annotations, which have been used to determine the degree of agreement.}
\label{tab:label-distribution}
\begin{small}
\begin{center}
\begin{tabular}{|l|C{2.5cm}|C{2.5cm}|C{2.5cm}|C{2.5cm}|}
\hline
 & \text{$\%$Negative} & \text{$\%$Neutral} & \text{$\%$Positive} & \text{Count} \\
\hline
Sentences with $100\%$ agreement& 13.4 & 61.4 & 25.2 & 2259 \\
Sentences with $>75\%$ agreement & 12.2 & 62.1 & 25.7 & 3448\\
Sentences with $>66\%$ agreement & 12.2 & 60.1 & 27.7 & 4211\\
Sentences with $>50\%$ agreement & 12.5 & 59.4 & 28.2 & 4840\\
\hline
\end{tabular}

\end{center}
\end{small}
\end{table*}

\section{Experiments and results}

The purpose of the LPS algorithm is to detect the semantic orientation of news phrases in a manner that is optimal in terms of accuracy, precision and recall. In this section, we construct a series of experiments to evaluate the performance of the model with respect to a number of baseline algorithms. As dataset for the experiments, we have used the financial phrase-bank described in the previous section. Example phrases with missing annotation labels and phrases that have been labeled by the annotators as inconsistent have been discarded from the evaluations. The system used in the experiments was implemented using Java-based software on top of  Stanford's CoreNLP framework,  which provides tools for standard document preprocessing tasks and extraction of phrase-structure information. The SVM models used within our LPS-algorithm and some of the baselines where learned using the LIBSVM implementation for maximum margin classification available in JavaML library. In all estimations the default parameters were used without optimizing the models for any specific criteria.

\subsection{Choice of benchmark algorithms}

To evaluate the benefits of the LPS algorithm, a number of baseline models were constructed. Since our LPS approach is essentially a hybrid of rule-based linguistic models and machine learning techniques, one of the objectives in the experiments is to understand how the added layers of rules contribute to the overall model performance when combined with a maximum margin classifier. Therefore, the benchmark algorithms featured below have been chosen to represent different levels of model complexity ranging from simple word based algorithms towards the model proposed by Moilanen et al.~\citeyear{moilanen10} and our approach:

\begin{itemize}

\item Wordcount MPQA (W-MPQA): In this approach, we calculate the number of positive and negative words in each sentence using the MPQA dictionary. The sentences are then labeled using the following rules: no polarized words = neutral; 2/3, or more, negative words = negative; 2/3, or more, positive words = positive.

\item Wordcount Loughran and McDonald~\citeyear{loughran11} (W-Loughran): We use a similar method as with Wordcount MPQA, but use the positive and negative word lists by Loughran and McDonald which is expected to yield better performance for financial and economic texts than the more general MPQA dictionary.

\item Quasi-compositional polarity sequence model with MPQA dictionary (SVM-MPQA): As the primary baseline in the experiments, we consider the polarity sequence model proposed by Moilanen et al.~\citeyear{moilanen10}. In the paper, they compared a number of alternative models with varying levels of complexity, but taken as a whole they found that a simple polarity sequence model with only POS tag level information outperformed their more complicated models relying on complete phrase structure information. The version considered here is their POS tag level model with MPQA dictionary as the source of polarity information.

\item LPS model without entity pruning rules (R-LPS): Finally, to show the relevance of entity pruning rules, we consider as a baseline a restricted version of the LPS. The model is equivalent to LPS in all other respects except for the use of entity pruning rules. The model has also access to the lexicon proposed in this paper.
\end{itemize}

\subsection{Performance comparison}\label{sec:experiment-1}

Tables~\ref{tab:data-1} and~\ref{tab:data-2} show performance of the models on the four reference datasets defined based on the phrase bank with different degrees of inter-annotator agreement. The results reported for the algorithms with a machine learning component (i.e. MPQA and the two variants of LPS) are computed using 10-fold cross-validation. 


\begin{table*}
\caption{Experimental results for sentences with strong majority agreement. Accuracy and F1 score are the most important metrics. All results are computed using 10-fold cross-validation.}\label{tab:data-1}
\vskip 0.2cm
\begin{footnotesize}
\begin{tabular}{|l|C{2.5cm}|C{2.5cm}|C{2.5cm}|C{2.5cm}|C{2.5cm}|}
\multicolumn{6}{l}{Panel A: Performance on sentences with $100\%$ annotator-agreement.} \\
\hline
&		\text{W-MPQA}	&	\text{W-Loughran}	&	\text{SVM-MPQA\tnote{a}}	&	\text{R-LPS}	&	\text{LPS} 		\\
\hline
\multicolumn{6}{|l|}{ Positive sentences} \\
\hline
Accuracy	&	0.659	&	0.755	&	0.746	&	0.858	&	\textbf{0.869}	\\
Recall	&	0.519	&	0.125	&	0.060	&	0.698	&	\textbf{0.737}	\\
Precision	&	0.374	&	0.563	&	0.466	&	0.728	&	\textbf{0.742}	\\
F1-score	&	0.435	&	0.204	&	0.106	&	0.713	&	\textbf{0.739}	\\
\hline
\multicolumn{6}{|l|}{ Neutral sentences} \\
\hline
Accuracy	&	0.586	&	0.625	&	0.652	&	\textbf{0.851}	&	0.828	\\
Recall	&	0.581	&	0.914	&	\textbf{0.963}	&	0.887	&	0.868	\\
Precision	&	0.694	&	0.635	&	0.645	&	\textbf{0.872}	&	0.854	\\
F1-score	&	0.632	&	0.750	&	0.773	&	\textbf{0.880}	&	0.861	\\
\hline
\multicolumn{6}{|l|}{ Negative sentences} \\
\hline
Accuracy	&	0.829	&	0.849	&	0.870	&	0.947	&	\textbf{0.951}	\\
Recall	&	0.370	&	0.162	&	0.205	&	\textbf{0.799}	&	0.789	\\
Precision	&	0.364	&	0.360	&	0.534	&	0.801	&	\textbf{0.839}	\\
F1-score	&	0.367	&	0.223	&	0.296	&	0.800	&	\textbf{0.813}	\\
\hline	
\multicolumn{6}{l}{} \\
\multicolumn{6}{l}{Panel B: Performance on sentences with $>75\%$ annotator-agreement.} \\
\hline
&		\text{W-MPQA}	&	\text{W-Loughran}	&	\text{SVM-MPQA\tnote{a}}	&	\text{R-LPS}	&	\text{LPS} 		\\
\hline

\multicolumn{6}{|l|}{ Positive sentences} \\
\hline
Accuracy	&	0.651	&	0.758	&	0.744	&	0.826	&	\textbf{0.836}	\\
Recall	&	0.557	&	0.166	&	0.077	&	0.614	&	\textbf{0.658}	\\
Precision	&	0.378	&	0.602	&	0.511	&	0.677	&	\textbf{0.690}	\\
F1-score	&	0.451	&	0.260	&	0.133	&	0.644	&	\textbf{0.674}	\\
\hline
\multicolumn{6}{|l|}{ Neutral sentences} \\
\hline
Accuracy	&	0.571	&	0.636	&	0.657	&	\textbf{0.799}	&	0.792	\\
Recall	&	0.553	&	0.904	&	\textbf{0.973}	&	0.865	&	0.837	\\
Precision	&	0.694	&	0.649	&	0.650	&	0.821	&	\textbf{0.830}	\\
F1-score	&	0.616	&	0.756	&	0.779	&	\textbf{0.842}	&	0.833	\\
\hline
\multicolumn{6}{|l|}{ Negative sentences} \\
\hline
Accuracy	&	0.841	&	0.863	&	0.886	&	0.939	&	\textbf{0.945}	\\
Recall	&	0.367	&	0.195	&	0.162	&	0.707	&	\textbf{0.800}	\\
Precision	&	0.354	&	0.378	&	0.630	&	\textbf{0.771}	&	0.760	\\
F1-score	&	0.360	&	0.257	&	0.258	&	0.738	&	\textbf{0.780}	\\
\hline
\end{tabular}
\end{footnotesize}
\end{table*}
\vskip 0.5cm


\begin{table*}
\caption{Experimental results for sentences with weak majority agreement. Accuracy and F1 score are the most important metrics. All results are computed using 10-fold cross-validation.}\label{tab:data-2}
\vskip 0.2cm
\begin{footnotesize}
\begin{tabular}{|l|C{2.5cm}|C{2.5cm}|C{2.5cm}|C{2.5cm}|C{2.5cm}|}
\multicolumn{6}{l}{Panel A: Performance on sentences with  $>66\%$ annotator-agreement} \\
\hline
&		\text{W-MPQA}	&	\text{W-Loughran}	&	\text{SVM-MPQA\tnote{a}}	&	\text{R-LPS}	&	\text{LPS} 		\\
\hline
\multicolumn{6}{|l|}{ Positive sentences} \\
\hline
Accuracy	&	0.642	&	0.741	&	0.722	&	\textbf{0.799}	&	0.798	\\
Recall	&	0.572	&	0.182	&	0.076	&	0.592	&	\textbf{0.816}	\\
Precision	&	0.399	&	0.606	&	0.494	&	\textbf{0.651}	&	0.599	\\
F1-score	&	0.470	&	0.279	&	0.132	&	0.620	&	\textbf{0.691}	\\
\hline
\multicolumn{6}{|l|}{ Neutral sentences} \\
\hline
Accuracy	&	0.563	&	0.622	&	0.632	&	\textbf{0.761}	&	0.753	\\
Recall	&	0.534	&	0.894	&	\textbf{0.966}	&	0.830	&	0.705	\\
Precision	&	0.672	&	0.631	&	0.626	&	0.786	&	\textbf{0.858}	\\
F1-score	&	0.595	&	0.740	&	0.759	&	\textbf{0.807}	&	0.774	\\
\hline
\multicolumn{6}{|l|}{ Negative sentences} \\
\hline
Accuracy	&	0.845	&	0.865	&	0.883	&	0.931	&	\textbf{0.937}	\\
Recall	&	0.379	&	0.214	&	0.142	&	0.681	&	\textbf{0.768}	\\
Precision	&	0.370	&	0.399	&	0.584	&	\textbf{0.731}	&	0.729	\\
F1-score	&	0.375	&	0.278	&	0.228	&	0.705	&	\textbf{0.748}	\\
\hline	
\multicolumn{6}{l}{} \\
\multicolumn{6}{l}{Panel B: Performance on sentences with $>50\%$ annotator-agreement (Gold standard).} \\
\hline
&		\text{W-MPQA}	&	\text{W-Loughran}	&	\text{SVM-MPQA\tnote{a}}	&	\text{R-LPS}	&	\text{LPS} 		\\
\hline

\multicolumn{6}{|l|}{ Positive sentences} \\
\hline
Accuracy	&	0.630	&	0.732	&	0.716	&	0.776	&	\textbf{0.786}	\\
Recall	&	\textbf{0.568}	&	0.187	&	0.068	&	0.500	&	0.535	\\
Precision	&	0.391	&	0.573	&	0.465	&	0.629	&	\textbf{0.645}	\\
F1-score	&	0.463	&	0.282	&	0.119	&	0.557	&	\textbf{0.585}	\\
\hline
\multicolumn{6}{|l|}{ Neutral sentences} \\
\hline
Accuracy	&	0.548	&	0.613	&	0.623	&	0.729	&	\textbf{0.735}	\\
Recall	&	0.516	&	0.882	&	\textbf{0.965}	&	0.836	&	0.809	\\
Precision	&	0.650	&	0.623	&	0.616	&	0.741	&	\textbf{0.760}	\\
F1-score	&	0.575	&	0.730	&	0.752	&	\textbf{0.786}	&	0.784	\\
\hline
\multicolumn{6}{|l|}{ Negative sentences} \\
\hline
Accuracy	&	0.848	&	0.863	&	0.880	&	0.922	&	\textbf{0.933}	\\
Recall	&	0.374	&	0.222	&	0.134	&	0.613	&	\textbf{0.772}	\\
Precision	&	0.388	&	0.407	&	0.574	&	\textbf{0.721}	&	0.716	\\
F1-score	&	0.381	&	0.287	&	0.217	&	0.662	&	\textbf{0.743}	\\
\hline
\end{tabular}
\end{footnotesize}
\end{table*}
\vskip 0.5cm


First, considering the results reported for datasets with strong majority agreement in Table~\ref{tab:data-1}, we find the performance quite encouraging. Both LPS and the reduced LPS models outperform the MPQA baseline and Wordcount-based voting rules in all sentence classes. The accuracy levels achieved by the better performing LPS algorithm ranged from 0.828 to 0.951 on sentences with 100\% agreement, and between 0.792 and 0.945 on the sentences with more than 75\% agreement. Also the F1 score, which is defined as the harmonic mean of precision and recall was very high for LPS in comparison to the corresponding ranges for MPQA, W-Loughran and W-MPQA. The performance gain of LPS over Reduced LPS in terms of F1 score and accuracy lends support for the use of pruning rules within the LPS model.

When comparing Table~\ref{tab:data-1} with the results for weaker majority agreement given in Table~\ref{tab:data-2}, we find that the performance is not greatly affected by the degree of inter-annotator agreement. Regardless of the reference dataset, the results produced by LPS and reduced LPS show clear gains over the other baseline algorithms. The overall performance of the models is only slightly weaker when compared to results reported in Table~\ref{tab:data-1}. This can be attributed to the increased number of positive sentences in the data, where annotator agreement on the correct label is weaker. For negative sentences, on the other hand, the performance is quite close to results obtained under strong majority agreement. Since the baseline models are ordered according to their complexity, the performance gains obtained by using more sophisticated lexicons and learning algorithms can be conveniently highlighted using W-MPQA as a reference point. While the use of better lexicons alone is sufficient to boost performance considerably, as indicated by the use of Loughran and McDonald's dictionary instead of the general MPQA dictionary, we find the role of good learning algorithms also important. 



\subsection{Sources of error in LPS models and directions for development}

According to the above experiments, the performance of the LPS models was quite strong regardless of the chosen gold standard. However, there is always room for improvement. To provide an idea on the sources of error and guide further development of the models to fix the shortcomings of the current models, we conducted an error analysis to identify the most common mistakes made by the LPS models. A random sample of 300 misclassified sentences was chosen, and we asked our Annotator A to review them based on the type of error that the algorithm is making. The results of the study are reported in Table~\ref{tab:lps-error-analysis}. In summary, the main sources of error turned out to be as follows: (i) the lack of knowledge on what is relevant for a company's success (i.e. ``inability to recognize significance of events" and ``need for more context"); (ii) inability to assess the credibility of a sentence (i.e. ``company talking in advertising like-tone about its own operations" or ``positive convention of talking about something"); and (iii) the failure to identify what is new information (e.g., detection of changes in numbers, time expressions and detection of roles).

\begin{table*}[htdp]
\caption{Sources of error in LPS model}
\begin{center}
\begin{tabular}{|l|C{3.5cm}|C{3.5cm}|}
\hline
\text{Error-type} & \text{$\%$ of errors} & \text{$\%$ as the only error} \\
\hline
Need for more context & 43.0 & 20.0 \\
Inability to recognize significance of events & 26.7 & 19.3 \\
Company talking in advertising like -tone about its' own & \multirow{2}{*}{10.0} & \multirow{2}{*}{3.3} \\
operations & & \\
Mismatched lexicon items & 9.0 & 7.7 \\
Polysemy of words and expressions & 8.7 & 0.7 \\
Positive convention of talking about something & 6.3 & 2.0 \\
Expressions missing from lexicon & 6.0 & 1.0 \\
Inability to understand magnitude and value of items & 5.3 & 0.3 \\
Inability to detect changes in numbers & 4.3 & 3.7 \\
Inability to detect roles in sentence & 4.3 & 0.3 \\
Errors in pruning rules and construction of entity sequences & 4.0 & 0.3 \\
Use of longer expressions and interpreting non-words as words & 4.0 & 0.0 \\
Inability to detect time expressions in the sentence & 3.3 & 2.7 \\
Borderline cases: sentences that could be tagged by & \multirow{2}{*}{2.7} & \multirow{2}{*}{1.0} \\
human as either/or & & \\
Sentences with multiple parts & 1.7 & 0.3 \\
Inability to reason from text & 1.3 & 0.7 \\
Wrong label in training data & 1.3 & 0.3 \\
\hline
\end{tabular}
\end{center}
\label{tab:lps-error-analysis}
\end{table*}


One of the key difficulties in phrase level sentiment analysis is the lack of deeper contextual information. Since each phrase has to be interpreted in isolation, a human reader would often prefer to see more context than one sentence in order to draw any conclusions. For instance, although acquisition events are commonly described in quite positive tone (e.g., ``The acquisition of Sampo Bank makes strategic sense for DB"), it is quite uncertain whether they will add or destroy value. Therefore, human analysts tend to be generally skeptical about the positiveness of such news, and would require a considerable amount of background information before wanting to judge statement as positive or negative. In addition to the lack of prior knowledge on the companies, the LPS algorithms are unable to distinguish between positiveness following from company's own statements vs. independent reviews about the companies. Also some events, such as nominations of new executives, are conventionally described in an overly positive manner instead of reflecting the actual facts. An interesting problem is also the detection of roles or perspective in a sentence (i.e. from whose viewpoint do we interpret the sentence when multiple parties are involved). Good news for someone may well be bad for another. For example, a sentence may talk about a company getting money from another company after a legal case. Clearly, there are still a number of ways to improve the performance of the sentiment models reported in this paper. In addition to the enrichment of lexicon with weights for different concepts and events, an important direction for future research will be to examine how phrase level models can be merged with content models~\cite{gabrilovich10,malo11,malo13}).

\section{Conclusions}

With the growing demand for sentiment analysis tools in financial and economic applications, it is increasingly important to pay attention to the ability of the models to capture the domain-specific use of language. As observed in the previous studies on general sentiment analysis, it is well-known that models that work in one domain may not work well in another one. In particular, when considering the specialized vocabulary encountered in finance and economics, building a model requires a combination of expert information in the form of high-quality lexicons as well as more sophisticated learning algorithms, which are better able to account for the contextual dependence of semantic orientations. The contributions in this paper can be summarized as follows. First, to support training and evaluation of the models, we have constructed a phrase-bank, which provides a comprehensive cross-section of sentences encountered in financial news and company press releases. The sentences were processed by a group of 16 annotators with an appropriate business education. The dataset is available to other researchers under a noncommercial license. The second contribution was to propose a technique for enhancing the financial lexicons with attributes, which help to identify expected direction of events that affect overall sentiment. Finally, we proposed a linearized phrase-structure model, which can effectively accommodate the influence of verbs and directional expressions in financial and economic domains. The performance of the proposed LPS model was compared with a number of baselines, and the results showed clear evidence on the benefits of using a machine learning approach in combination with domain-knowledge on financial entities.


\bibliographystyle{apacite}

\end{document}